\PassOptionsToPackage{dvipsnames}{xcolor}
\documentclass[sigconf, natbib=true, anonymous=false, nonacm]{acmart}
\AtBeginDocument{%
  }
\setcopyright{acmlicensed}
\copyrightyear{2026}
\acmYear{2026}
\setcopyright{acmlicensed}
\acmConference[SIGIR '26]{49th International ACM SIGIR Conference on Research and Development in Information Retrieval}{July 20--24, 2026}{Melbourne, Australia}
\acmISBN{978-1-4503-XXXX-X/2018/06}

\usepackage{enumitem}
\usepackage{multirow}
\usepackage{subcaption}
\usepackage{pgfplots}
\pgfplotsset{compat=1.18}

\begin{document}

\title{%
Individual Turing Test: A Case Study of LLM-based Simulation Using Longitudinal Personal Data
}

\newcommand{\ziyi}[1]{\textcolor{Violet}{#1}}


\author{Minghao Guo}
\affiliation{%
  \institution{Rutgers University}
  \city{New Brunswick}
  \state{NJ}
  \country{USA}}
\email{minghao.guo@rutgers.edu}

\author{Ziyi Ye}
\authornote{Corresponding author.}
\affiliation{%
  \institution{Fudan University}
  \city{Shanghai}
  \country{China}}
\email{zyye@fudan.edu.cn}

\author{Wujiang Xu}
\affiliation{%
  \institution{Rutgers University}
  \city{New Brunswick}
  \state{NJ}
  \country{USA}}
\email{wujiang.xu@rutgers.edu}

\author{Xi Zhu}
\affiliation{%
  \institution{Rutgers University}
  \city{New Brunswick}
  \state{NJ}
  \country{USA}}
\email{xi.zhu@rutgers.edu}

\author{Wenyue Hua}
\affiliation{%
  \institution{Microsoft}
  \city{New York}
  \state{NY}
  \country{USA}}
\email{wenyuehua@microsoft.com}

\author{Dimitris N. Metaxas}
\affiliation{%
  \institution{Rutgers University}
  \city{New Brunswick}
  \state{NJ}
  \country{USA}}
  \email{dnm@cs.rutgers.edu}

\renewcommand{\shortauthors}{Trovato et al.}

\begin{abstract}
Large Language Models~(LLMs) have demonstrated remarkable human-like capabilities, yet their ability to replicate a specific individual remains under-explored.
This paper presents a case study to investigate LLM-based individual simulation with a volunteer-contributed archive of private messaging history spanning over ten years.
Based on the messaging data, we propose the ``Individual Turing Test'' to evaluate whether acquaintances of the volunteer can correctly identify which response in a multi-candidate pool most plausibly comes from the volunteer.
We investigate prevalent LLM-based individual simulation approaches including: fine-tuning, retrieval-augmented generation~(RAG), memory-based approach, and hybrid methods that integrate fine-tuning and RAG or memory.
Empirical results show that current LLM-based simulation methods do not pass the Individual Turing Test, but they perform substantially better when the same test is conducted on strangers to the target individual.
Additionally, while fine-tuning improves the simulation in daily chats representing the language style of the individual, retrieval-augmented and memory-based approaches demonstrate stronger performance on questions involving personal opinions and preferences.
These findings reveal a fundamental trade-off between parametric and non-parametric approaches to individual simulation with LLMs when given a longitudinal context.

\end{abstract}




\settopmatter{printacmref=false} %
\renewcommand\footnotetextcopyrightpermission[1]{} %
\pagestyle{plain} %

\maketitle

\begin{table*}[t!]
    \centering
    \caption{Case Study: Qualitative Comparison of Responses to \textit{``Are you more drawn to the opportunities in big cities, or the slower pace of life in smaller cities?''}. Avg. Rank denotes the mean ranking position assigned by human judges (lower is better).}
    \label{tab:case_study}
    \begin{tabular}{l p{13cm} c}
        \toprule
        \textbf{Method} & \textbf{Response Content} & \textbf{Avg. Rank} \\
        \midrule
        \textbf{Ground Truth} & \textit{Big cities. You get to meet all kinds of people and be exposed to a lot more.} & \textbf{2.00} \\
        \midrule
        \textbf{A-Mem+LoRA} & \textit{Small cities are too slow for me. I can’t really live like that. I’d rather go out and work elsewhere.} & \textbf{2.33} \\
        \midrule
        \textbf{RAG+LoRA} & \textit{Probably the opportunities. Big cities just offer more possibilities.} & 3.17 \\
        \midrule
        \textbf{LoRA} & \textit{Ah, that’s a tough one. Both have their own appeal! What about you?} & 3.33 \\
        \midrule
        \textbf{RAG+Base} & \textit{I do envy how many opportunities big cities have. Do you like the quiet of smaller towns? Of course, competition in big cities is intense too. What about you?} & 3.50 \\
        \midrule
        \textbf{A-Mem+Base} & \textit{I lean toward the opportunities in big cities---there’s simply more. But sometimes I also want to slow down and enjoy the quiet of a small town.} & 3.67 \\
        \bottomrule
    \end{tabular}
\vspace{-0.25cm}
\end{table*}

\section{Introduction}

Large Language Models (LLMs) have recently demonstrated remarkable capabilities in individual simulation, leveraging their strong language understanding, implicit persona learning, and reasoning abilities~\cite{park2023generativeagentsinteractivesimulacra, shao2023characterllmtrainableagentroleplaying, jin2024impact}. 
Existing individual-level simulation with LLMs primarily relies on either \emph{parametric} (e.g., fine-tuning model weights) or \emph{non-parametric} (e.g., retrieval-augmented prompting that injects external evidence at inference time) methods~\cite{jin2025exploring, jin2025disentangling, mei2025r,guo2026deepsieveinformationsievingllmasaknowledgerouter,lin2025cachemechanismagentrag,li-etal-2025-cardiverse,wei2025mlpmemoryretrieverpretrainedmemory,wang2026ragrouter}. 
Recent agent-based frameworks further extend these techniques to support interactive behaviors and longer-term memory mechanisms~\cite{zheng2020personalizeddialoguegenerationdiversified, wang2025recursivelysummarizingenableslongterm, luo2025code}.

Despite these advances, most existing studies on LLM-based individual simulation are conducted under limited temporal scope, typically conditioning on static user profiles or short conversational context ~\cite{du2025twinvoicemultidimensionalbenchmarkdigital,du2025understandingeffectopinionpolarization,du-etal-2025-simvbg}.
Crucially, longitudinal personal data encodes multiple behavioral dimensions, including stylistic patterns in daily interactions (e.g., brevity, emojis, habitual phrasing) as well as personal opinions and preferences that can be inferred from the long-term context.
On the other hand, existing evaluations largely rely on automated language similaritymetrics or judgments from other LLMs, which are insufficient for assessing whether machine-generated responses are truly indistinguishable from a target individual’s authentic expressions for human judges~\cite{li2024llmsasjudgescomprehensivesurveyllmbased, gehrmann2021gem, clark-etal-2021-thats, hua2024trustagent, ji2025moralbench, lin2024battleagent, hua2025disentangling, zeng2025uncertainty}.

To address this gap, we propose \emph{Individual Turing Test}, which aims to evaluate whether machine-generated responses can be indistinguishable from the authentic expressions of a \emph{specific individual} with the individual's longitudinal context.
Unlike conventional Turing test~\cite{turing2007computing, kiela2021dynabenchrethinkingbenchmarkingnlp} that focuses on general human-likeness, our test requires the LLMs to learn the target individual's language style and personal opinions.
The test is evaluated on the individual's acquaintances and compared with a \emph{General Turing Test} that is evaluated on strangers~(strangers are given the individual's basic profile). 
This evaluation enables us to assess the LLMs' ability to generate responses that are indistinguishable from humans with different levels of knowledge about the individual.

We conduct the evaluation based on a longitudinal dataset of the volunteer's private messaging history spanning over ten years and compared four prevalent individual simulation approaches: fine-tuning, retrieval-augmented generation~(RAG), memory-based approach, and hybrid methods that integrate two of them.
A qualitative example is shown in Table~\ref{tab:case_study}, where the average rank of the ground truth response is lower than the other responses.
This suggests that current LLMs are unable to convincingly simulate the individual from an acquaintances' perspective. 
However, evidence from the General Turing Test suggests that LLMs can succeed when evaluated on strangers of the individual. Furthermore, we identify a style-substance trade-off: parametric fine-tuning captures linguistic style, whereas non-parametric retrieval mechanisms ensure opinion consistency.
Additionally, we find that simulation fidelity is sensitive to the temporal scope of historical context, highlighting the necessity of longitudinal data.

In summary, this work makes three contributions.
(1) We propose the \textbf{Individual Turing Test}, a protocol for evaluating individual simulation, which reveals a gap between general human-likeness and identity-level authenticity.
(2) We compare \textbf{parametric}, \textbf{non-parametric}, and \textbf{hybrid} personalization strategies, revealing a trade-off between stylistic alignment and opinion grounding.
(3) We demonstrate that \textbf{temporal scope} affects performance, with gains from recent history saturating beyond a threshold.

\section{Methodology}

\subsection{Task Definition}
The \emph{individual Turing test} aims to generate responses that are indistinguishable from those of a specific target individual under realistic conversational contexts.
Given a target individual $u$ and a set of conversational prompts $\mathcal{Q}$, the target individual produces a ground-truth response $r^{\text{gt}}(q)$ for each $q \in \mathcal{Q}$.
A simulation method $m$ generates a corresponding response $r^{(m)}(q)$ conditioned on the same prompt $q$.
Then we evaluate whether the simulated responses can faithfully reflect the target individual's authentic behavior under both rigorous human judgment and automated evaluation metrics.

\subsection{Dataset Construction}
To enable individual-specific and long-term evaluation, we conduct a study based on a longitudinal private messaging archive contributed by a volunteer.
Ethical approval was acquired via an anonymous organization.
The dataset spans approximately ten years of daily conversations, covering a wide range of casual and personal interaction scenarios.
All data were anonymized and processed to remove sensitive personal information.
Conversational instances are constructed with a two-stage time-window heuristic: consecutive messages from the same speaker are merged within 2 minutes, and user--assistant turns are paired using a 5-minute window.
To further improve data quality, we apply overlap-based heuristics (substring/equality checks, and $>0.8$ character-coverage overlap for short replies) together with a lightweight blacklist to prune repetitive, low-information responses (e.g., acknowledgments and media placeholders).
Overall, the dataset contains 12{,}151 conversations, 72{,}652 total messages, and 1{,}157{,}842 training tokens.
Notably, the scale of this dataset is substantially larger than those used in prior studies, which enables a more comprehensive evaluation of generalization and long-term consistency in individual simulation.



\subsection{Evaluation Protocol}
We adopt a \emph{hybrid} evaluation framework that combines \textbf{human judgment} with \textbf{automated language similarity metrics} across two representative prompt types: \textbf{daily conversations} and \textbf{personal opinions}, to comprehensively assess the overall fidelity and the underlying behavioral mechanisms of individual simulation.

\paragraph{Human judgment: Individual Turing Test.}
We formulate the \emph{Individual Turing Test} as a ranking-based identification task.
For each prompt $q \in \mathcal{Q}$, evaluators are presented with a shuffled pool containing (i) the ground-truth response $r^{\text{gt}}(q)$ from the target individual and (ii) simulated responses $\{r^{(m)}(q)\}_{m \in \mathcal{M}}$ generated by different individual simulation approaches.
Evaluators then rank all responses based on how plausibly they correspond to the target individual (lower is more plausible).
The evaluators include 7 acquaintances of the individual who are familiar with the individual with at least 3 years of friendship, but explicitly excluding immediate relatives of the individual.

In addition, we run an auxiliary \emph{General Turing Test} to probe general capability under the same human-judgment protocol. 
In the General Turing Test, responses are evaluated by 5 strangers who are provided with the target individual's basic profile, including working experience, education, and demographic information.

Based on the ranking data, we compute the selection rate (SR), defined as the proportion of times a method is ranked first, which reflects the most confident identification. For each method $m$, we compute its \emph{selection rate}
\[
\text{SR}(m) = \frac{C_m}{\sum_{m' \in \mathcal{M} \cup \{\text{gt}\}} C_{m'}} ,
\]
where $C_m$ denotes the number of times responses generated by method $m$ are selected by human evaluators.
Across methods, lower selection rates than ground truth indicate a clear remaining distinguishability gap under human judgment, while closer selection rates suggest stronger individual-level authenticity.

\paragraph{Automatic language similarity metrics.}
In addition to the human judgment, we also calculate automatic metrics including BLEU-1/2, ROUGE-L, Precision, Recall, and Distinct-2 to reflect lexical accuracy, structural, content coverage and lexical diversity.


\paragraph{Prompt types.}
We consider two representative prompt types that differ in their temporal and
semantic focus, with 30 prompts sampled for each category.
\textbf{Daily conversations} capture routine, situational interactions that
primarily reflect immediate context and timeliness, aligned with everyday
chit-chat and open-domain dialogue settings commonly studied in prior work
(e.g., \citep{jandaghi-etal-2024-faithful,zhang2018personalizingdialogueagentsi}), such as
\textit{``What news did you see recently?''}.
In contrast, \textbf{Personal opinions} probe longer-term preferences or judgment tendencies that are expected to remain consistent over time, which follows opinion- and preference-oriented prompts in personalization and user modeling studies (e.g., \citep{Inglehart2014wvs,zhang2018personalizingdialogueagentsi,ye2022towards}), such as
\textit{``When reading news, do you care more about facts or opinions?''}.

\section{Experiment Results and Discussion}

We evaluate LLM-based individual simulation from three perspectives.
First, we compare performance under the Individual Turing Test (judged by acquaintances) and the General Turing Test (judged by strangers), using the latter as a baseline for the capability of LLM in general human-likeness.
Second, we examine how different simulation strategies perform under the Individual Turing Test.
Finally, we analyze the impact of temporal scope of the individual's messaging history on the performance of the LLMs.

\paragraph{Simulation Methods.}
To study how different individual simulation approaches affect the performance of LLMs,
we systematically compare parametric and non-parametric approaches.
We adopt LoRA-7B~\cite{hu2021loralowrankadaptationlarge} as the parametric setting,
as full fine-tuning on longitudinal data was found to cause
substantial performance degradation.
For non-parametric modeling, we consider both retrieval-based
RAG~\cite{lewis2021retrievalaugmentedgenerationknowledgeintensivenlp}
and memory-based A-Mem~\cite{xu2025mem}.
We further evaluate hybrid variants that combine LoRA with RAG or A-Mem.

\paragraph{Implementation Details.}
We select Qwen2.5-7B as the simulation backbone LLM for all methods.
For LoRA, we fine-tuned the model on 2 $\times$ A800 GPUs with rank $r=4$ and dropout 0.3, targeting projection modules ($q\_\text{proj}$, $v\_\text{proj}$). 
The model was trained for 2 epochs with learning rate $1\times 10^{-4}$ and a cosine schedule. 
For RAG/A-Mem/Hybrid inference, we retrieved top-$k=5$ memories using BGE-M3 embeddings with cosine-similarity filtering ($\text{min\_cosine}=0.35$) and de-duplication ($\text{dedup\_cosine}=0.92$).
We applied controlled decoding for inference with temperature 0.85, repetition penalty 1.2, and max tokens 80.


\paragraph{Statistical Protocol.} To rigorously assess performance differences, we conducted \textit{post-hoc} pairwise comparisons based on rank differences. Unless otherwise stated, significance was determined using one-sided binomial tests.

\subsection{Turing Test: Capability of LLM-Based Individual Simulation}

\begin{figure}[t]
\centering

\def\Methods{LoRA,RAG+Base,A-Mem+Base,RAG+LoRA,A-Mem+LoRA,{Ground\ Truth}}

\begin{tikzpicture}
\def\GTTTotalVotes{120} 

\def\VGRAGBase{23}
\def\VGAMemBase{54}
\def\VGLoRA{4}
\def\VGRAGLoRA{3}
\def\VGAMemLoRA{8} 
\def\VGGT{9}

\pgfmathsetmacro{\SRAGBase}{100*\VGRAGBase/\GTTTotalVotes}
\pgfmathsetmacro{\SAMemBase}{100*\VGAMemBase/\GTTTotalVotes}
\pgfmathsetmacro{\SLoRA}{100*\VGLoRA/\GTTTotalVotes}
\pgfmathsetmacro{\SRAGLoRA}{100*\VGRAGLoRA/\GTTTotalVotes}
\pgfmathsetmacro{\SAMemLoRA}{100*\VGAMemLoRA/\GTTTotalVotes}
\pgfmathsetmacro{\SGT}{100*\VGGT/\GTTTotalVotes}

\pgfmathsetmacro{\POShareLoRA}{40}
\pgfmathsetmacro{\POShareRAGBase}{67}
\pgfmathsetmacro{\POShareAMemBase}{70}
\pgfmathsetmacro{\POShareRAGLoRA}{30}
\pgfmathsetmacro{\POShareAMemLoRA}{45}
\pgfmathsetmacro{\POShareGT}{49}

\pgfmathsetmacro{\SPOLOra}{\SLoRA*\POShareLoRA/100}
\pgfmathsetmacro{\SPORAGBase}{\SRAGBase*\POShareRAGBase/100}
\pgfmathsetmacro{\SPOAMemBase}{\SAMemBase*\POShareAMemBase/100}
\pgfmathsetmacro{\SPORAGLoRA}{\SRAGLoRA*\POShareRAGLoRA/100}
\pgfmathsetmacro{\SPOAMemLoRA}{\SAMemLoRA*\POShareAMemLoRA/100}
\pgfmathsetmacro{\SPOGT}{\SGT*\POShareGT/100}

\pgfmathsetmacro{\SDailyLoRA}{\SLoRA-\SPOLOra}
\pgfmathsetmacro{\SDailyRAGBase}{\SRAGBase-\SPORAGBase}
\pgfmathsetmacro{\SDailyAMemBase}{\SAMemBase-\SPOAMemBase}
\pgfmathsetmacro{\SDailyRAGLoRA}{\SRAGLoRA-\SPORAGLoRA}
\pgfmathsetmacro{\SDailyAMemLoRA}{\SAMemLoRA-\SPOAMemLoRA}
\pgfmathsetmacro{\SDailyGT}{\SGT-\SPOGT}

\begin{axis}[
    ybar stacked,
    width=\columnwidth,
    height=3.7cm,
    ymin=0, ymax=55,
    bar width=7pt,
    enlarge x limits=0.14,
    ylabel={Selection Rate (\%)},
    symbolic x coords={LoRA-only,RAG+Base,A-Mem+Base,RAG+LoRA,A-Mem+LoRA,{Ground\ Truth}},
    xtick=data,
    x tick label style={rotate=25, anchor=east, font=\scriptsize},
    yticklabel style={font=\scriptsize},
    ylabel style={font=\scriptsize},
    legend style={font=\scriptsize, at={(0.02,0.98)}, anchor=north west, draw=none},
    legend cell align=left,
]
\addplot coordinates {
    (LoRA-only,\SDailyLoRA)
    (RAG+Base,\SDailyRAGBase)
    (A-Mem+Base,\SDailyAMemBase)
    (RAG+LoRA,\SDailyRAGLoRA)
    (A-Mem+LoRA,\SDailyAMemLoRA)
    ({Ground\ Truth},\SDailyGT)
};
\addlegendentry{Daily conversations}

\addplot coordinates {
    (LoRA-only,\SPOLOra)
    (RAG+Base,\SPORAGBase)
    (A-Mem+Base,\SPOAMemBase)
    (RAG+LoRA,\SPORAGLoRA)
    (A-Mem+LoRA,\SPOAMemLoRA)
    ({Ground\ Truth},\SPOGT)
};
\addlegendentry{Personal opinions}
\end{axis}
\end{tikzpicture}

\vspace{0.15cm}
{\small\textbf{(a) General Turing Test} (strangers, $N=5$).}

\vspace{0.35cm}

\begin{tikzpicture}
\def\TotRAGBase{7.7}
\def\TotAMemBase{8.1}
\def\TotLoRAonly{11.3}
\def\TotRAGLoRA{17.1}
\def\TotAMemLoRA{17.6}
\def\TotGT{32.5}

\def\TOneRAGBase{2.8}
\def\TOneAMemBase{2.0}
\def\TOneLoRAonly{8.9}
\def\TOneRAGLoRA{8.7}
\def\TOneAMemLoRA{7.7}
\def\TOneGT{14.2}

\pgfmathsetmacro{\TTwoRAGBase}{\TotRAGBase-\TOneRAGBase}
\pgfmathsetmacro{\TTwoAMemBase}{\TotAMemBase-\TOneAMemBase}
\pgfmathsetmacro{\TTwoLoRAonly}{\TotLoRAonly-\TOneLoRAonly}
\pgfmathsetmacro{\TTwoRAGLoRA}{\TotRAGLoRA-\TOneRAGLoRA}
\pgfmathsetmacro{\TTwoAMemLoRA}{\TotAMemLoRA-\TOneAMemLoRA}
\pgfmathsetmacro{\TTwoGT}{\TotGT-\TOneGT}

\begin{axis}[
    ybar stacked,
    width=\columnwidth,
    height=3.7cm,
    ymin=0, ymax=35,
    bar width=7pt,
    enlarge x limits=0.14,
    ylabel={Selection Rate (\%)},
    symbolic x coords={LoRA-only,RAG+Base,A-Mem+Base,RAG+LoRA,A-Mem+LoRA,{Ground\ Truth}},
    xtick=data,
    x tick label style={rotate=25, anchor=east, font=\scriptsize},
    yticklabel style={font=\scriptsize},
    ylabel style={font=\scriptsize},
    legend style={font=\scriptsize, at={(0.02,0.98)}, anchor=north west, draw=none},
    legend cell align=left,
]
\addplot coordinates {
    (LoRA-only,\TOneLoRAonly)
    (RAG+Base,\TOneRAGBase)
    (A-Mem+Base,\TOneAMemBase)
    (RAG+LoRA,\TOneRAGLoRA)
    (A-Mem+LoRA,\TOneAMemLoRA)
    ({Ground\ Truth},\TOneGT)
};
\addlegendentry{Daily conversations}

\addplot coordinates {
    (LoRA-only,\TTwoLoRAonly)
    (RAG+Base,\TTwoRAGBase)
    (A-Mem+Base,\TTwoAMemBase)
    (RAG+LoRA,\TTwoRAGLoRA)
    (A-Mem+LoRA,\TTwoAMemLoRA)
    ({Ground\ Truth},\TTwoGT)
};
\addlegendentry{Personal opinions}
\end{axis}
\end{tikzpicture}

\vspace{0.15cm}
{\small\textbf{(b) Individual Turing Test} (acquaintances, $N=7$).}

\vspace{-0.15cm}
\caption{
\textbf{Comparison between General and Individual Turing Tests.}
We report selection rates for each method under two evaluation settings: 
(a) \textit{General Turing Test}, where responses are judged by strangers, and 
(b) \textit{Individual Turing Test}, where responses are judged by acquaintances of the target individual. 
Results are stacked by prompt type, distinguishing \textit{daily conversations} and \textit{personal opinion} prompts. 
}

\label{fig:tt_compare}
\vspace{-0.25cm}
\end{figure}

We first conduct a comparative analysis of LLM performance under the \textit{General Turing Test}
and the \textit{Individual Turing Test}, as shown in Figure~\ref{fig:tt_compare}. 
Under the General Turing Test, ground-truth responses do not
achieve the highest selection rate. 
Instead, LLM-based simulation methods~(e.g., A-Mem+Base) achieve comparable or even higher selection rates, suggesting that LLM-based simulation can produce responses that appear plausibly
human when evaluated by strangers.
For example, post-hoc pairwise comparisons show that the A-Mem+Base responses significantly perform better than the ground truth with $p=4.79\times 10^{-8}$.

In contrast, under the Individual Turing Test, the ground-truth responses achieve the highest selection rate.
Post-hoc pairwise comparisons confirm that the ground-truth responses are significantly better than the best simulation method A-Mem+LoRA with $p=6.21\times 10^{-3}$.
This indicates that current LLMs still struggle to faithfully replicate the target individual's behavior from the  perspective of close acquaintances. This divergence between the two evaluation settings suggests that general human-likeness does not imply identity-specific fidelity. 

\subsection{Comparison of Different Simulation Methods}
\paragraph{Performance Hierarchy.} Overall, as shown in Figure~\ref{fig:tt_compare}b, the hybrid methods achieve the strongest performance, with LoRA-only approaches following, while retrieval-only and memory-only approaches perform the worst.
Pairwise comparisons confirm this hierarchy:
RAG+LoRA significantly outperforms RAG+Base
(win rate $=0.694$, $p=2.88\times 10^{-11}$),
LoRA-only also significantly surpasses RAG+Base
(win rate $=0.608$, $p=5.59\times 10^{-4}$),
and RAG+LoRA further exceeds LoRA-only
(win rate $=0.614$, $p=5.72\times 10^{-5}$).
Together, these results establish a clear and consistent performance hierarchy:\textbf{
\textit{Hybrid} $>$ \textit{Parametric only} $>$ \textit{Non-parametric only}}. 

\paragraph{Mechanistic Differences Across Prompt Types.}
To understand this hierarchy, we analyze performance across prompt categories.
Parametric adaptation (LoRA) contributes primarily to style-dominated daily interactions, where matching conversational tone, brevity, and informal phrasing is critical.
In contrast, non-parametric retrieval and memory mechanisms (RAG and A-Mem) contribute more substantially to prompts involving personal opinions and preferences, where consistency with historically grounded opinions is essential.
In other words, parametric methods primarily dictate the \emph{style},
while non-parametric memory secures the \emph{substance}, and hybrid models unify both strengths to achieve stronger performance across two prompt types.

We illustrate this divergence with a concrete example.
When asked \textit{``What kinds of movies or TV shows do you usually like?''},
the LoRA-only model produces a short, stylistically aligned reply
such as \textit{``I’m into mysteries and sci-fi. What about you?''}.
In contrast, RAG+Base retrieves specific historical references
(e.g., \textit{``I mostly like action and fantasy—I’m really into stuff like
\emph{The Lord of the Rings}.''}),
but the phrasing may sound less individualized.
The hybrid RAG+LoRA combines stylistic alignment and grounded detail:
\textit{``Mostly sci-fi, like \emph{Interstellar}. How about you?''}.
This qualitative contrast mirrors the quantitative patterns
observed in Figure~\ref{fig:tt_compare}b.


\paragraph{Alignment with Automated Metrics.}
These human-judgment findings are further reflected in
automated metrics (Table~\ref{tab:automated_metrics}).
BLEU-1 measures unigram overlap and therefore reflects
token-level lexical alignment with the target individual,
while BLEU-2 and ROUGE-L capture local and sequence-level
structural consistency.
Precision penalizes excessive or off-distribution generation
by measuring the proportion of generated tokens that appear
in the reference responses,
whereas Recall rewards broader coverage of historically
grounded content.
Distinct-2 quantifies bigram diversity and is sensitive to
mode collapse.
Under this lens, parametric adaptation tends to improve
BLEU-1, Precision, and Distinct-2,
indicating stronger stylistic constraint and controlled diversity.
In contrast, retrieval and memory mechanisms contribute more
substantially to Recall and lexical coverage metrics,
reflecting their role in grounding responses in historical content.
Hybrid models balance both effects,
achieving the strongest overall metric profiles.

\begin{table}[t]
\centering
\small
\caption{\textbf{Selected automated metrics (diagnostic).} \textbf{Hybrid} methods (LoRA + retrieval/\allowbreak memory) consistently outperform \textbf{LoRA-only} and \textbf{retrieval/\allowbreak memory-only} baselines.}
\label{tab:automated_metrics}
\setlength{\tabcolsep}{3.5pt}
\resizebox{\columnwidth}{!}{%
\begin{tabular}{l|cccccc}
\toprule
\textbf{Method} & \textbf{BLEU-1} $\uparrow$ & \textbf{BLEU-2} $\uparrow$ & \textbf{ROUGE-L} $\uparrow$ & \textbf{Prec.} $\uparrow$ & \textbf{Rec.} $\uparrow$ & \textbf{Dist-2} $\uparrow$ \\
\midrule
LoRA-only & 0.1578 & 0.0501 & \textbf{0.1496} & 0.1565 & 0.2128 & 0.7171 \\
RAG+Base & 0.1011 & 0.0264 & 0.1089 & 0.1005 & 0.2949 & 0.6407 \\
A-Mem+Base & 0.1113 & 0.0309 & 0.1228 & 0.1109 & \textbf{0.3613} & 0.6433 \\
RAG+LoRA & 0.1837 & 0.0735 & 0.1456 & \textbf{0.1776} & 0.1714 & 0.7680 \\
A-Mem+LoRA & \textbf{0.1983} & \textbf{0.1175} & 0.1457 & 0.1649 & 0.1586 & \textbf{0.7881} \\
\bottomrule
\end{tabular}%
}
\vspace{-0.25cm}
\end{table}

\subsection{Impact of Temporal Scope of Messaging History}

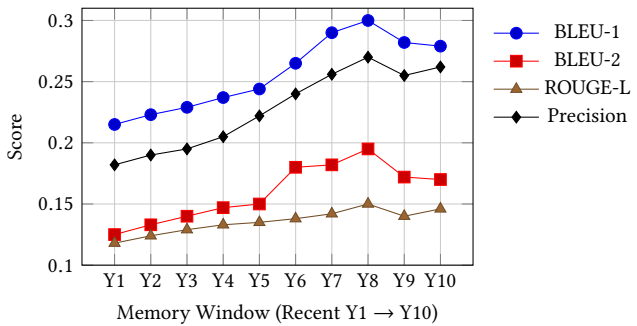
\begin{figure}[t]
\centering
\begin{tikzpicture}
\begin{axis}[
    width=0.8\columnwidth,
    height=5.0cm,
    ymin=0.10, ymax=0.31,
    ymajorgrids=true,
    xmajorgrids=true,
    grid style={gray!40},
    xlabel={Memory Window (Recent Y1 $\rightarrow$ Y10)},
    ylabel={Score},
    symbolic x coords={Y1,Y2,Y3,Y4,Y5,Y6,Y7,Y8,Y9,Y10},
    xtick=data,
    tick label style={font=\small},
    label style={font=\small},
    legend style={font=\small, at={(1.03,0.98)}, anchor=north west, draw=none},
]
\addplot+[mark=*, mark size=2.2pt] coordinates
{(Y1,0.215) (Y2,0.223) (Y3,0.229) (Y4,0.237) (Y5,0.244) (Y6,0.265) (Y7,0.290) (Y8,0.300) (Y9,0.282) (Y10,0.279)};
\addlegendentry{BLEU-1}
\addplot+[mark=square*, mark size=2.2pt] coordinates
{(Y1,0.125) (Y2,0.133) (Y3,0.140) (Y4,0.147) (Y5,0.150) (Y6,0.180) (Y7,0.182) (Y8,0.195) (Y9,0.172) (Y10,0.170)};
\addlegendentry{BLEU-2}
\addplot+[mark=triangle*, mark size=2.4pt] coordinates
{(Y1,0.118) (Y2,0.124) (Y3,0.129) (Y4,0.133) (Y5,0.135) (Y6,0.138) (Y7,0.142) (Y8,0.150) (Y9,0.140) (Y10,0.146)};
\addlegendentry{ROUGE-L}
\addplot+[mark=diamond*, mark size=2.2pt] coordinates
{(Y1,0.182) (Y2,0.190) (Y3,0.195) (Y4,0.205) (Y5,0.222) (Y6,0.240) (Y7,0.256) (Y8,0.270) (Y9,0.255) (Y10,0.262)};
\addlegendentry{Precision}
\end{axis}
\end{tikzpicture}
\caption{Effect of expanding recent memory on hybrid simulation (A-Mem + LoRA).
$Y_i$ denotes using the most recent $i$ years of dialogue history
(e.g., $Y_1$ = most recent year only; $Y_{10}$ = most recent ten years).}
\label{fig:memory_window}
\vspace{-0.25cm}
\end{figure}

We further analyze how the temporal scope of the individual's messaging history
affects the performance of the LLMs.
Using the strongest hybrid configuration (A-Mem + LoRA),
we expand the temporal scope of the messaging history from only the most recent year to the most recent ten years, and evaluate the performance of the LLMs using the automated language similarity metrics.
Figure~\ref{fig:memory_window} shows that performance improves consistently as additional recent years are incorporated, with steady gains observed up to approximately eight years of dialogue history.
BLEU-1 and Precision increase monotonically,
indicating stronger token-level consistency and stylistic alignment,
while BLEU-2 and ROUGE-L also exhibit gradual improvements,
suggesting enhanced lexical grounding.

However, when extending beyond roughly eight years,
performance plateaus and slightly declines.
Including more distant history may introduce outdated or less temporally relevant patterns, which can dilute signals from more recent behavioral context.
These results suggest that effective personalization depends
not only on the amount of memory incorporated,
but also on temporal relevance.
Expanding recent history improves grounding up to a threshold,
after which additional memory coverage yields only minimal further performance improvement.

\section{Conclusion}

We investigated LLMs for individual simulation through both the General and Individual Turing Tests. While hybrid methods (parametric adaptation combined with retrieval or memory) consistently outperform single-component approaches, all models remain substantially below human performance, revealing a persistent authenticity gap. Notably, the divergence between the two evaluation settings highlights that general human-likeness does not guarantee identity-level fidelity under acquaintance judgment.

Our findings suggest a structural division of labor in individual simulation:
parametric adaptation primarily captures stylistic regularities,
whereas non-parametric memory mechanisms preserve opinion consistency
and historically grounded content.
Moreover, effective personalization depends not only on memory coverage
but also on temporal relevance, as expanding recent history improves
performance only up to a saturation threshold.

These results indicate that faithful individual simulation requires coordinating style, substance, and temporal alignment. Beyond improving raw generation quality, future research should further investigate how to dynamically balance parametric adaptation and memory grounding under different conversational contexts. Future work may also explore more principled memory selection strategies and recency-aware architectures to further narrow the authenticity gap and better capture identity-level behavioral signals.

\bibliographystyle{ACM-Reference-Format}
\bibliography{sample-base}

@article{ji2025moralbench,
  title={Moralbench: Moral evaluation of llms},
  author={Ji, Jianchao and Chen, Yutong and Jin, Mingyu and Xu, Wujiang and Hua, Wenyue and Zhang, Yongfeng},
  journal={ACM SIGKDD Explorations Newsletter},
  volume={27},
  number={1},
  pages={62--71},
  year={2025},
  publisher={ACM New York, NY, USA}
}

@inproceedings{zeng2025uncertainty,
  title={Uncertainty Quantification for Multiple-Choice Questions is Just One-Token Deep},
  author={Zeng, Qingcheng and Jin, Mingyu and Yu, Qinkai and Wang, Zhenting and Hua, Wenyue and Sun, Guangyan and Meng, Yanda and Ma, Shiqing and Wang, Qifan and Juefei-Xu, Felix and others},
  booktitle={Proceedings of the 34th ACM International Conference on Information and Knowledge Management},
  pages={5474--5478},
  year={2025}
}

@article{luo2025code,
  title={Code agent can be an end-to-end system hacker: Benchmarking real-world threats of computer-use agent},
  author={Luo, Weidi and Zhang, Qiming and Lu, Tianyu and Liu, Xiaogeng and Hu, Bin and Chiu, Hung-Chun and Ma, Siyuan and Zhang, Yizhe and Xiao, Xusheng and Cao, Yinzhi and others},
  journal={arXiv preprint arXiv:2510.06607},
  year={2025}
}

@inproceedings{hua2025disentangling,
  title={Disentangling logic: The role of context in large language model reasoning capabilities},
  author={Hua, Wenyue and Zhu, Kaijie and Li, Lingyao and Fan, Lizhou and Jin, Mingyu and Lin, Shuhang and Xue, Haochen and Li, Zelong and Wang, JinDong and Zhang, Yongfeng},
  booktitle={Findings of the Association for Computational Linguistics: ACL 2025},
  pages={19219--19242},
  year={2025}
}

@inproceedings{lin2024battleagent,
  title={Battleagent: Multi-modal dynamic emulation on historical battles to complement historical analysis},
  author={Lin, Shuhang and Hua, Wenyue and Li, Lingyao and Chang, Che-Jui and Fan, Lizhou and Ji, Jianchao and Hua, Hang and Jin, Mingyu and Luo, Jiebo and Zhang, Yongfeng},
  booktitle={Proceedings of the 2024 Conference on Empirical Methods in Natural Language Processing: System Demonstrations},
  pages={172--181},
  year={2024}
}

@inproceedings{hua2024trustagent,
  title={Trustagent: Towards safe and trustworthy llm-based agents},
  author={Hua, Wenyue and Yang, Xianjun and Jin, Mingyu and Li, Zelong and Cheng, Wei and Tang, Ruixiang and Zhang, Yongfeng},
  booktitle={Findings of the Association for Computational Linguistics: EMNLP 2024},
  pages={10000--10016},
  year={2024}
}

@inproceedings{jin2025disentangling,
  title={Disentangling memory and reasoning ability in large language models},
  author={Jin, Mingyu and Luo, Weidi and Cheng, Sitao and Wang, Xinyi and Hua, Wenyue and Tang, Ruixiang and Wang, William Yang and Zhang, Yongfeng},
  booktitle={Proceedings of the 63rd Annual Meeting of the Association for Computational Linguistics (Volume 1: Long Papers)},
  pages={1681--1701},
  year={2025}
}

@inproceedings{jin2025exploring,
  title={Exploring concept depth: How large language models acquire knowledge and concept at different layers?},
  author={Jin, Mingyu and Yu, Qinkai and Huang, Jingyuan and Zeng, Qingcheng and Wang, Zhenting and Hua, Wenyue and Zhao, Haiyan and Mei, Kai and Meng, Yanda and Ding, Kaize and others},
  booktitle={Proceedings of the 31st international conference on computational linguistics},
  pages={558--573},
  year={2025}
}

@inproceedings{ye2022towards,
  title={Towards a better understanding of human reading comprehension with brain signals},
  author={Ye, Ziyi and Xie, Xiaohui and Liu, Yiqun and Wang, Zhihong and Chen, Xuesong and Zhang, Min and Ma, Shaoping},
  booktitle={Proceedings of the ACM Web Conference 2022},
  pages={380--391},
  year={2022}
}

@inproceedings{jin2024impact,
  title={The impact of reasoning step length on large language models},
  author={Jin, Mingyu and Yu, Qinkai and Shu, Dong and Zhao, Haiyan and Hua, Wenyue and Meng, Yanda and Zhang, Yongfeng and Du, Mengnan},
  booktitle={Findings of the Association for Computational Linguistics: ACL 2024},
  pages={1830--1842},
  year={2024}
}

@misc{du2025twinvoicemultidimensionalbenchmarkdigital,
      title={TwinVoice: A Multi-dimensional Benchmark Towards Digital Twins via LLM Persona Simulation}, 
      author={Bangde Du and Minghao Guo and Songming He and Ziyi Ye and Xi Zhu and Weihang Su and Shuqi Zhu and Yujia Zhou and Yongfeng Zhang and Qingyao Ai and Yiqun Liu},
      year={2025},
      eprint={2510.25536},
      archivePrefix={arXiv},
      primaryClass={cs.CL},
      url={https://arxiv.org/abs/2510.25536}, 
}

@misc{guo2026deepsieveinformationsievingllmasaknowledgerouter,
      title={DeepSieve: Information Sieving via LLM-as-a-Knowledge-Router}, 
      author={Minghao Guo and Qingcheng Zeng and Xujiang Zhao and Yanchi Liu and Wenchao Yu and Mengnan Du and Haifeng Chen and Wei Cheng},
      year={2026},
      eprint={2507.22050},
      archivePrefix={arXiv},
      primaryClass={cs.CL},
      url={https://arxiv.org/abs/2507.22050}, 
}

@misc{park2023generativeagentsinteractivesimulacra,
      title={Generative Agents: Interactive Simulacra of Human Behavior}, 
      author={Joon Sung Park and Joseph C. O'Brien and Carrie J. Cai and Meredith Ringel Morris and Percy Liang and Michael S. Bernstein},
      year={2023},
      eprint={2304.03442},
      archivePrefix={arXiv},
      primaryClass={cs.HC},
      url={https://arxiv.org/abs/2304.03442}, 
}

@misc{shao2023characterllmtrainableagentroleplaying,
      title={Character-LLM: A Trainable Agent for Role-Playing}, 
      author={Yunfan Shao and Linyang Li and Junqi Dai and Xipeng Qiu},
      year={2023},
      eprint={2310.10158},
      archivePrefix={arXiv},
      primaryClass={cs.CL},
      url={https://arxiv.org/abs/2310.10158}, 
}

@misc{zheng2020personalizeddialoguegenerationdiversified,
      title={Personalized Dialogue Generation with Diversified Traits}, 
      author={Yinhe Zheng and Guanyi Chen and Minlie Huang and Song Liu and Xuan Zhu},
      year={2020},
      eprint={1901.09672},
      archivePrefix={arXiv},
      primaryClass={cs.CL},
      url={https://arxiv.org/abs/1901.09672}, 
}

@misc{wang2025recursivelysummarizingenableslongterm,
      title={Recursively Summarizing Enables Long-Term Dialogue Memory in Large Language Models}, 
      author={Qingyue Wang and Yanhe Fu and Yanan Cao and Shuai Wang and Zhiliang Tian and Liang Ding},
      year={2025},
      eprint={2308.15022},
      archivePrefix={arXiv},
      primaryClass={cs.CL},
      url={https://arxiv.org/abs/2308.15022}, 
}

@misc{li2024llmsasjudgescomprehensivesurveyllmbased,
      title={LLMs-as-Judges: A Comprehensive Survey on LLM-based Evaluation Methods}, 
      author={Haitao Li and Qian Dong and Junjie Chen and Huixue Su and Yujia Zhou and Qingyao Ai and Ziyi Ye and Yiqun Liu},
      year={2024},
      eprint={2412.05579},
      archivePrefix={arXiv},
      primaryClass={cs.CL},
      url={https://arxiv.org/abs/2412.05579}, 
}

@inproceedings{gehrmann2021gem,
  title={The gem benchmark: Natural language generation, its evaluation and metrics},
  author={Gehrmann, Sebastian and Adewumi, Tosin and Aggarwal, Karmanya and Ammanamanchi, Pawan Sasanka and Aremu, Anuoluwapo and Bosselut, Antoine and Chandu, Khyathi Raghavi and Clinciu, Miruna and Das, Dipanjan and Dhole, Kaustubh and others},
  booktitle={Proceedings of the 1st Workshop on Natural Language Generation, Evaluation, and Metrics (GEM 2021)},
  pages={96--120},
  year={2021}
}

@inproceedings{clark-etal-2021-thats,
    title = "All That{'}s `Human' Is Not Gold: Evaluating Human Evaluation of Generated Text",
    author = "Clark, Elizabeth  and
      August, Tal  and
      Serrano, Sofia  and
      Haduong, Nikita  and
      Gururangan, Suchin  and
      Smith, Noah A.",
    editor = "Zong, Chengqing  and
      Xia, Fei  and
      Li, Wenjie  and
      Navigli, Roberto",
    booktitle = "Proceedings of the 59th Annual Meeting of the Association for Computational Linguistics and the 11th International Joint Conference on Natural Language Processing (Volume 1: Long Papers)",
    month = aug,
    year = "2021",
    address = "Online",
    publisher = "Association for Computational Linguistics",
    url = "https://aclanthology.org/2021.acl-long.565/",
    doi = "10.18653/v1/2021.acl-long.565",
    pages = "7282--7296"
}

@incollection{turing2007computing,
  title={Computing machinery and intelligence},
  author={Turing, Alan M},
  booktitle={Parsing the Turing test: Philosophical and methodological issues in the quest for the thinking computer},
  pages={23--65},
  year={2007},
  publisher={Springer}
}

@misc{kiela2021dynabenchrethinkingbenchmarkingnlp,
      title={Dynabench: Rethinking Benchmarking in NLP}, 
      author={Douwe Kiela and Max Bartolo and Yixin Nie and Divyansh Kaushik and Atticus Geiger and Zhengxuan Wu and Bertie Vidgen and Grusha Prasad and Amanpreet Singh and Pratik Ringshia and Zhiyi Ma and Tristan Thrush and Sebastian Riedel and Zeerak Waseem and Pontus Stenetorp and Robin Jia and Mohit Bansal and Christopher Potts and Adina Williams},
      year={2021},
      eprint={2104.14337},
      archivePrefix={arXiv},
      primaryClass={cs.CL},
      url={https://arxiv.org/abs/2104.14337}, 
}

@misc{wei2025mlpmemoryretrieverpretrainedmemory,
      title={MLP Memory: A Retriever-Pretrained Memory for Large Language Models}, 
      author={Rubin Wei and Jiaqi Cao and Jiarui Wang and Jushi Kai and Qipeng Guo and Bowen Zhou and Zhouhan Lin},
      year={2025},
      eprint={2508.01832},
      archivePrefix={arXiv},
      primaryClass={cs.CL},
      url={https://arxiv.org/abs/2508.01832}, 
}

@inproceedings{jandaghi-etal-2024-faithful,
    title = "Faithful Persona-based Conversational Dataset Generation with Large Language Models",
    author = "Jandaghi, Pegah  and
      Sheng, Xianghai  and
      Bai, Xinyi  and
      Pujara, Jay  and
      Sidahmed, Hakim",
    editor = "Nouri, Elnaz  and
      Rastogi, Abhinav  and
      Spithourakis, Georgios  and
      Liu, Bing  and
      Chen, Yun-Nung  and
      Li, Yu  and
      Albalak, Alon  and
      Wakaki, Hiromi  and
      Papangelis, Alexandros",
    booktitle = "Proceedings of the 6th Workshop on NLP for Conversational AI (NLP4ConvAI 2024)",
    month = aug,
    year = "2024",
    address = "Bangkok, Thailand",
    publisher = "Association for Computational Linguistics",
    url = "https://aclanthology.org/2024.nlp4convai-1.8/",
    pages = "114--139"
}

@article{mei2025r,
  title={R-wom: Retrieval-augmented world model for computer-use agents},
  author={Mei, Kai and Guo, Jiang and Chang, Shuaichen and Dong, Mingwen and Lee, Dongkyu and Niu, Xing and Jiang, Jiarong},
  journal={arXiv preprint arXiv:2510.11892},
  year={2025}
}

@misc{zhang2018personalizingdialogueagentsi,
      title={Personalizing Dialogue Agents: I have a dog, do you have pets too?}, 
      author={Saizheng Zhang and Emily Dinan and Jack Urbanek and Arthur Szlam and Douwe Kiela and Jason Weston},
      year={2018},
      eprint={1801.07243},
      archivePrefix={arXiv},
      primaryClass={cs.AI},
      url={https://arxiv.org/abs/1801.07243}, 
}

@electronic{Inglehart2014wvs,
  address = {Madrid},
  editor = {Inglehart, R. and Haerpfer, C. and Moreno, A. and Welzel, C. and Kizilova, K. and Diez-Medrano, J. and Lagos, M. and Norris, P. and Ponarin, E. and Puranen, B. and et al.},
  institution = {JD Systems Institute},
  title = {World Values Survey: Round Six - Country-Pooled Datafile Version},
  url = {www.worldvaluessurvey.org/WVSDocumentationWV6.jsp},
  year = 2014
}

@article{xu2025mem,
  title={A-Mem: Agentic memory for LLM agents},
  author={Xu, Wujiang and Liang, Zujie and Mei, Kai and Gao, Hang and Tan, Juntao and Zhang, Yongfeng},
  journal={arXiv preprint arXiv:2502.12110},
  year={2025}
}

@misc{lewis2021retrievalaugmentedgenerationknowledgeintensivenlp,
      title={Retrieval-Augmented Generation for Knowledge-Intensive NLP Tasks}, 
      author={Patrick Lewis and Ethan Perez and Aleksandra Piktus and Fabio Petroni and Vladimir Karpukhin and Naman Goyal and Heinrich Küttler and Mike Lewis and Wen-tau Yih and Tim Rocktäschel and Sebastian Riedel and Douwe Kiela},
      year={2021},
      eprint={2005.11401},
      archivePrefix={arXiv},
      primaryClass={cs.CL},
      url={https://arxiv.org/abs/2005.11401}, 
}

@misc{hu2021loralowrankadaptationlarge,
      title={LoRA: Low-Rank Adaptation of Large Language Models}, 
      author={Edward J. Hu and Yelong Shen and Phillip Wallis and Zeyuan Allen-Zhu and Yuanzhi Li and Shean Wang and Lu Wang and Weizhu Chen},
      year={2021},
      eprint={2106.09685},
      archivePrefix={arXiv},
      primaryClass={cs.CL},
      url={https://arxiv.org/abs/2106.09685}, 
}

@misc{lin2025cachemechanismagentrag,
      title={Cache Mechanism for Agent RAG Systems}, 
      author={Shuhang Lin and Zhencan Peng and Lingyao Li and Xiao Lin and Xi Zhu and Yongfeng Zhang},
      year={2025},
      eprint={2511.02919},
      archivePrefix={arXiv},
      primaryClass={cs.CL},
      url={https://arxiv.org/abs/2511.02919}, 
}

@misc{du2025understandingeffectopinionpolarization,
      title={Understanding the Effect of Opinion Polarization in Short Video Browsing}, 
      author={Bangde Du and Ziyi Ye and Monika Jankowska and Zhijing Wu and Qingyao Ai and Yiqun Liu},
      year={2025},
      eprint={2403.04184},
      archivePrefix={arXiv},
      primaryClass={cs.SI},
      url={https://arxiv.org/abs/2403.04184}, 
}

@inproceedings{du-etal-2025-simvbg,
    title = "{S}im{VBG}: Simulating Individual Values by Backstory Generation",
    author = "Du, Bangde  and
      Ye, Ziyi  and
      Wu, Zhijing  and
      Jankowska, Monika A.  and
      Zhu, Shuqi  and
      Ai, Qingyao  and
      Zhou, Yujia  and
      Liu, Yiqun",
    editor = "Christodoulopoulos, Christos  and
      Chakraborty, Tanmoy  and
      Rose, Carolyn  and
      Peng, Violet",
    booktitle = "Proceedings of the 2025 Conference on Empirical Methods in Natural Language Processing",
    month = nov,
    year = "2025",
    address = "Suzhou, China",
    publisher = "Association for Computational Linguistics",
    url = "https://aclanthology.org/2025.emnlp-main.662/",
    doi = "10.18653/v1/2025.emnlp-main.662",
    pages = "13093--13122",
    ISBN = "979-8-89176-332-6"
}

@inproceedings{li-etal-2025-cardiverse,
    title = "Cardiverse: Harnessing {LLM}s for Novel Card Game Prototyping",
    author = "Li, Danrui  and
      Zhang, Sen  and
      Sohn, Samuel S.  and
      Hu, Kaidong  and
      Usman, Muhammad  and
      Kapadia, Mubbasir",
    editor = "Christodoulopoulos, Christos  and
      Chakraborty, Tanmoy  and
      Rose, Carolyn  and
      Peng, Violet",
    booktitle = "Proceedings of the 2025 Conference on Empirical Methods in Natural Language Processing",
    month = nov,
    year = "2025",
    address = "Suzhou, China",
    publisher = "Association for Computational Linguistics",
    url = "https://aclanthology.org/2025.emnlp-main.1511/",
    doi = "10.18653/v1/2025.emnlp-main.1511",
    pages = "29735--29762",
    ISBN = "979-8-89176-332-6"
}

@article{wang2026ragrouter,
  title={RAGRouter-Bench: A Dataset and Benchmark for Adaptive RAG Routing},
  author={Wang, Ziqi and Zhu, Xi and Lin, Shuhang and Xue, Haochen and Guo, Minghao and Zhang, Yongfeng},
  journal={arXiv preprint arXiv:2602.00296},
  year={2026}
}


\end{document}